# FROM *ALIFE* AGENTS TO A KINGDOM OF *N* QUEENS

HAN JING

*Dept. of Computer Sci., Univ. of Sci.&Tech. of China, Hefei, 230027,*
*P.R..China*
*E-mail: han_jing@263.net*

JIMING LIU

*Dept. of Computer Sci., Hong Kong Baptist University, Kowloon Tong*
*Hong Kong*
*E-mail: jiming@comp.hkbu.edu.hk*

CAI QINGSHENG

*Dept. of Computer Sci., Univ. of Sci.&Tech. of Chine, Hefei, 230027,*
*P.R.China*
*E-mail: qscai@ustc.edu.cn*

This paper presents a new approach to solving *N*-queen problems, which involves a model of distributed autonomous agents with artificial life (ALife) and a method of representing *N*-queen constraints in an agent environment. The distributed agents locally interact with their living environment, i.e., a chessboard, and execute their reactive behaviors by applying their behavioral rules for randomized motion, least-conflict position searching, and cooperating with other agents etc. The agent-based *N*-queen problem solving system evolves through selection and contest according to the rule of *Survival of the Fittest*, in which some agents will die or be eaten if their moving strategies are less efficient than others. The experimental results have shown that this system is capable of solving large-scale *N*-queen problems. This paper also provides a model of ALife agents for solving *general* CSPs.

## 1   Introduction

A Constraint Satisfaction Problem (CSP) consists of three components: variables, values, and constraints. The goal of solving a CSP is to find an assignment of values to variables such that all constraints are satisfied. Backtrack searching and its improvements [1-4] are commonly applied to solve CSPs. Suffering from the exponential growth in computing time, backtrack searching techniques are not a good candidate for solving large-scale CSPs. *N*-queen problem is a famous benchmark CSP, which requires to find the places for *n* indistinguishable queens on an $N \times N$ chessboard so that no two queens are placed within the same row, the same column, or the same diagonal. There exist solutions for the *N*-queen problem with *n* greater than or equal to 4 [3]. Backtrack searching techniques may systematically generate all possible solutions [1]. But in the worst case, backtrack searching is exponential in time. Stone [4] used various search heuristics to solve the *N*-queen problems for *n* up to 96. Kale [5] gave a new backtracking heuristic, which is capable of finding a solution in time proportional to $O(n^2)$. Later, Sosic and Gu [6] gave an efficient local search technique that runs in polynomial or liner time.

However, none of those algorithms considers how to solve the *m*-queen problem by re-using the information in solving an *N*-queen problem ($m \geq N$). The approach introduced here, an Artificial Life (ALife)-based agent CSP solver, utilizes evolutionary autonomous agents that can move, combine, and cease to exist during the course of interacting with a constraint conflict environment, according to the rule of *Survival of the Fittest*. The most distinct characteristics of our approach lie in that it is environment-driven, decentralized, and distributed in nature and relies on local agent "processes", whose behaviors are both easy to define and implement.

ALife-based evolutionary computation is concerned with applying the computational models of evolutionary processes to either achieving intelligent agent behaviors or solving real-life computation–intensive engineering problems [7-9]. Using ALife techniques as an approach to solving *N*-queen problems is a newly explored area of ALife research. At the same time, it also represents a new direction for CSPs studies.

The remainder of this paper is organized as follows: Section II states the key ideas behind our ALife-based agent CSP solver. Section III provides a model of the proposed autonomous agents, their reactive behaviors, and an ALife-based algorithm for evolving solutions for an *N*-queen problem. Section IV describes several computational experiments for investigating the key parameters of the agent-based CSP solving. Section V offers an ALife-based model for solving general CSPs. Finally, Section VI concludes the paper by highlighting the key contributions of this work.

## 2   The Key Ideas behind an ALife-Based Agent CSP Solver

In this paper, we will examine the use of simple distributed agents in solving CSPs. We have built an evolutionary agent system based on the rule of *Survival of the Fittest*, in which the characteristics of distributed computation and emergence are demonstrated. Here are the basic ideas underlying the development of an ALife-based agent CSP solver:

(a) Intelligence is emerged from many interacting individuals (agents) that make the system flexible and easy to evolve.
(b) Each agent can be very simple, with simple behavioral rules, and it is

autonomous.
(c) The complexity of the system lies in the number, the interrelation, and the diversity of individual agents.
(d) The competition for survival, reproduction, and mutation in individual agents are the causes of agent systems improvement.
(e) Besides providing contents to individual agents, an environment also serves as a medium for interaction among agents. So the agent environment is playing an essential role to the system.
(f) The agent system should be designed, and hence evolved, on the edge between chaos and order. There will be no creative power in a highly ordered system. On the other hand, it is hard to control the system and make it convergent if the system is too chaotic — the most difficult thing to implement.

The above ideas are reflected in our ALife-based agent CSP solver for tackling the benchmark problem of CSPs, *N*-queen problem. Although *N*-queen problem is not a distributed problem, it is possible to think of the chessboard as the agent environment and therefore model the *N*-queen problem into an ALife System. Our experimental results, as to be presented later, show that such an ALife system can efficiently find solutions to the *N*-queen problem and provide insights into agents for solving other CSPs.

## 3 How to Solve It?

There are several important notions in the simulations of nature: a collection of individuals, a living space for individuals, and a schedule of activities for those individuals. In our present work, the individuals are the agents, the living space is the environment in which the agents inhabit, and the schedule is the arrow of time along which the agents evolve. The relationships among them may be written as follows: **ALife System** :={Space, Schedule, AgentGroup}; **Schedule** :={timetable to update the changes and dispatch agents' actions}; **Space** :={physical properties and the positions of agents}; **AgentGroup** :={autonomous agent with some behavioral rules}.

### 3.1 Modeling an N-queen Problem

3.1.1 The chessboard is the environment (*space*).

**Figure 1.** Number of collisions.

Autonomous agents operate in an $N \times N$ square lattice. Each lattice records the number of collisions among the agents in the position of the lattice. Fig.1 shows an example of 4-Queen problem. Each numbers on the lattice denotes the number of collisions, and the circle indicates the existence of an agent. In addition, *space* dynamically tracks the positions of the agents. So *space* is a variable and will be instantly updated while agents are moving.

3.1.2    Queens are regarded as agents. More than one agent can be placed within one row. Following policies will be adopted by the agents:

(a) Each agent has its **Initial Energy** and some parameters for its behavioral rules;
(b) Agent will **die** when its energy is less than a threshold (suppose to be *zero*). According to the *Survival of the Fittest* principle, the poor-performance agent should be washed out; meanwhile, the possibility of finding a solution will be decreased if many agents are in the same row with different moving strategies.
(c) Agents can move on the chessboard ($N \times N$ lattice), but they can only **move to the right or the left**. This policy tries to avoid that the system becomes too chaotic and unable to reach convergence, and simplify the design and rules of agents.
(d) Agent will **lose energy** when it moves to a certain position. *StepLoseEnergy* is used to denote the lost energy at each movement. *While agent $\phi$ moves to a lattice where collision number is m, it will lose m units of energy. Meanwhile, agents who attack $\phi$ will also lose 1 unit of energy.* — This point is important because it is the only inherent drive for the rambling agents to find a solution. It will force the agent to move to the solution position because agent will lose no energy when it is in that position. It should think more about moving to the positions with fewer collisions to avoid losing energy.
(e) When agent $\phi$ moves to a lattice where other agents have already been, $\phi$ will compare its energy with the energy of all the other agents $\phi'$ that are in the same position. If the energy-difference is larger than the threshold of merge, ($\phi$. Energy - $\phi'$. Energy>*MergeThreshold*), $\phi$ will **eat** $\phi'$, and absorb the energy of $\phi'$. This kind of behavior can be called *Law of the Jungle*. Obviously, energy plays an important role for agents — 'more energy, less probability to be eaten'. This will encourage agents to try to avoid losing energy and select a good position whenever and wherever it moves. This policy will award energy to the agent with fit behavioral rules. Those agents that can survive due to their good strategies for moving are left in the system. That is what the system wants. It can be envisioned that when the system is running, more bad-strategy agents will vanish and the remainders are relatively better ones, meaning that the system is improved. However, the threshold value for merge should not set to be low, or eating will happen soon after a few movements, which may not distinguish the good strategies from the bad ones. On the other hand, if the threshold value is too high, few times of merge will happen for bad-strategy agents. Also, a high threshold will affect the speed of systems convergence, since there are too many agents alive over a long period.

(f) The simple **moving strategies** of agent are:
  i. **Randomized-move**. Move randomly with probability of *Random-p*. Randomized-move is necessary because without randomized movement the system will get into local optima and lose the chance for finding a solution. In addition, behaviors of human or other intelligent life forms sometimes are random.
  ii. **Least-Move**. Move to the position, which has the least collisions, with probability of *Least-P*. This strategy is instinctive for agents because moving to the least-collisions-value position will lose least energy.
  iii. **Coop-Move**. Move to a position, which will not attack agents in some special rows, with probability of *Coop-P*. This policy attempts to promote *Cooperation of Multi-Agents*. Here, *CoopVector*, a vector, specifies in what rows agents should not be attacked while moving. The vector is an *N*-bit-length string with '1' and '0'. See Fig.2, *V* is the *CoopVector* of agent $\phi$ in (2,1), *V*[1]=0, *V*[3]=1, *V*[4]=1. $\phi$ will try to avoid moving to those positions that will attack two agents in row 3 and one agent in row 4. Sometimes the agent may not be able to find a position that can satisfy the whole vector.

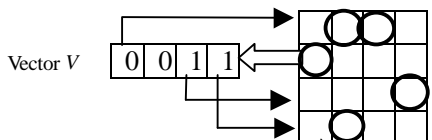 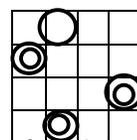

**Figure 2.** Vector of $\phi$.     **Figure 3**. A solution state.

3.1.3 In the initialization of the system, *m* $(1 \leq m < N)$ agents are randomly placed in each row.

More than one agent in the same row will bring more difficulty for system to find a solution. On the other hand, it provides chance for the system to wash out some bad-strategy agents. Moreover, if strategies of two agents are similar, they will often move to the same position — *Great Minds Think Alike*. This phenomenon is often observed in our practical experiments. In the following presentation, *m* is denoted by *RowNum*, $\phi_0$ is used to specify the initial agent group.

### 3.1.4 A solution

At time *t*, if there is only one position per row occupied by agents, and queens in those agent-occupy positions will not attack each other, a solution is found. Fig.3 shows a solution for 4-queen problem, and two circles mean two agents in that lattice. So it is possible to find a solution even when move than one agent inside one row. Those agents in the same position can be regarded as one queen. Good-strategy agents inside one row often have the similar strategies and the similar movements. They are *in the same camp* and seldom disturb the solution forming. The practical experiments show that nearly all the solutions have more than *N* agents when the initial *RowNum*>1, there are some agents overlapping in some lattices.

After a solution is found, the current agent group $\Phi$ can be utilized to find the next solution. Since $\Phi$ has learnt something, its performance is surely better than that of $\Phi_0$. Difference between $\Phi$ and $\Phi_0$ lies in two ways: There exists $\phi \in \Phi_0$, but $\phi \notin \Phi$; energy of each agent has changed since they lost or absorbed energy in the process. To utilize $\Phi$ to find another solution is a way of re-using the information in solving *N*-queen problem.

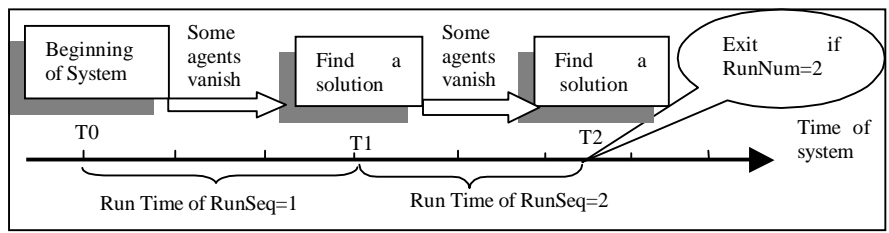

**Figure 4**. RunSeq and RunNm.

### *3.2 ALife-Based Algorithm for an N-Queen Problem*

Based on the model mentioned above, the ALife-based algorithm for *N*-queen is presented in Fig. 5. *RunSeq* records how many times the current agent group continuously finds solutions, and *RunNum* tells the system how many times we expect the system to find solution. See Fig.4, after system initialization at time T0, the initial agent group $\Phi$ will lose some agents while running. At time T1, a solution S1 is found and (T1-T0) is called '*runtime for finding solution when RunSeq=1*' or '*runtime for the first time*'. Suppose *RunNum=2*, the system will use the survived agents to find another solution. So (T2-T1) is called runtime when *RunSeq=2*, or runtime for the second time. *No-solution-state* means there exists $j \in [1…N]$, all agents in row *j* vanished. *MaxRandom-p*, *MaxLeast-P* and *MaxCoop-P* set the maximal values of *Random-p*, *Least-P* and *Coop-P* for all agents.

1. Initialization of system. Set the parameters:
    *RunSeq* = 0; *RowNum*= how many agents are placed on each row;
    /* System parameters for agents */
    Initialize (*InitialEnergy*, *StepLoseEnergy*, *MaxRandom-p*, *MaxLeast-P*, *MaxCoop-P*, *MergeThreshold*);
2. Initialize the *Space*; Initialize agents and add to agent group Φ; *Timer* = 0;
3. Randomly place agents to its corresponding row ; *RunSeq*++;
4. **While not** (*current-space = no-solution-state*) **do**
    **For** all $\phi \in \Phi$ **do** /* schedule in time t (t=*timer*) */
      Compute the possibilities of each strategy,
         and select one behavior among (Random-Move, Least-Move, Coop-Move);
      Compute the new position (*x', y*);
      **If** current-position(*x,y*)=(*x',y*) **then** stay
      **else if** (Other agent $\phi'$ in position(*x', y*)) **then**  //* move to other position */
        **if** |$\phi$.Energy-$\phi'$.Energy | > *MergeThreshold* **then** $\phi$ (or $\phi'$) eat $\phi'$ (or $\phi$);
        Update $\phi$.energy; Update energy of all agents that are attacked by $\phi$. Update *Space*;
        **If** *current-space = solution-state* **GoTo** 6
      *Timer* ++;
5. Output "Fail to find a solution!" ; **GoTo** 2; /* when current space = no-solution-state */
6. Output solution; **If** *RunSeq<= RunNum* **then  GoTo** 3;

**Figure 5.** ALife-based algorithm for an *N*-queen problem. Space complexity is $O(N^2)$.

## 4 Experimental Results

Having presented the model of ALife-based agent CSP solving, in what follows we describe some typical results of experimental validation. These results have indicated that the proposed approach can solve large-scale *N*-queen problems within a reasonable time. **Up to now, our agent system can solve 1500-queen in few hundreds of seconds with a computing environment of CPU: P233, RAM:32M, and OS:Win95.** Fig.6 shows the runtime result of solving the *N*-queen problem where *N* changes from *N=100* to *N=1500* with some specific parameters. Additional observations of several quantitative comparisons of different parameters are presented in the next subsections.

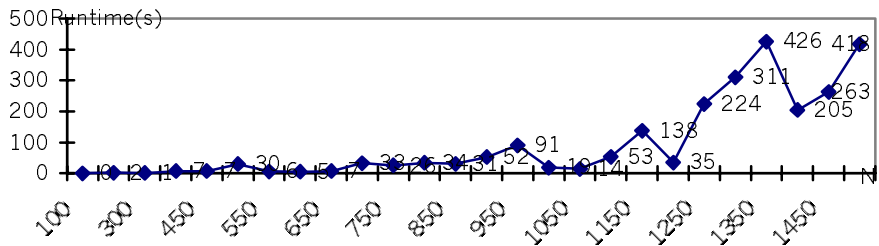

**Figure 6.** Runtime of N=(100, 1500) (Δ=10).

*4.1 RowNum, the initial number of agents per row*—Survival of the Fittest

**Experiment 1:** N=(20,410),RowNum=1,RunNum=1,MaxRandom-p=1,MaxLeast-P=90, MaxCoop-P=20
**Experiment 2:** N=(20,410),RowNum=2,RunNum=2,MaxRandom-p=1,MaxLeast-P=90, MaxCoop-P=20
**Experiment 3:** N=(20,410),RowNum=3,RunNum=3,MaxRandom-p=1, MaxLeast-P=90,MaxCoop-P=20

In Experiment 1, RowNum=1 indicates that agents are actually queens. System will fail to find a solution if any agent vanishes and has no chance to wash out any bad-strategy agents. But, there are no moving opinion conflicts inside one row, whereas in Experiments 2 and 3 agents inside one row will sometimes move to different positions. However, this provides a chance for systems optimization according to the rule of *Survival of the Fittest* and awards for those with good strategies.

Of great interest is Table 1. One conclusion can be drawn is that the longer the system evolves, the better its efficiency will become. In each experiment:
  **(runtime of RunSeq=3)<(runtime of RunSeq=2)<(runtime of RunSeq=1).**

From Fig.7(a), it can be found that the advantage of RowNum=3, RunSeq=3 becomes more evident when *N* is increasing. Fig.7(b) shows that RowNum=3 is still the best one though it has the largest population of agents.

Table 1. Efficiency measures in 3 experiments

| Row Num | Run Seq | Averaged Runtime | Linear-regress analysis of Runtime | Death number | Conclusion of runtime |
|---|---|---|---|---|---|
| 1 | 1 | 150.0 | 1.696*$N$-214.48 | Zero | Long when N>200, not stable. |
| 2 | 1 | 54.5 | 0.497*$N$-52.15 | $\approx TN$ | Between RowNum=1 and RowNum=3. |
|   | 2 | 18.5 | 0.176*$N$-19.47 | A few | Much better than RunSeq=1, more stable. |
| 3 | 1 | 46.0 | 0.772*$N$-110.1 | $\approx TN$ | Shortest, most stable among RunSeq =1 |
|   | 2 | 7.6 | 0.125*$N$-18.52 | A few | Much better than RunSeq=1, more stable. |
|   | 3 | 6.0 | 0.154*$N$-28.72 | $\approx TN$ | A little better than RunSeq =2, more stable. It is the best curve among all. |

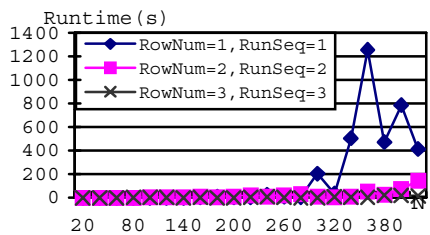
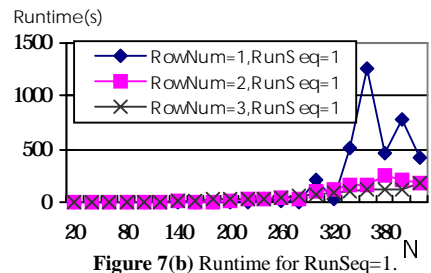

**Figure 7(a)** Best runtime curves for each experiment.   **Figure 7(b)** Runtime for RunSeq=1.

**Observation 1:** *System becomes more efficient and stable after washing out agents according to the rule of Survival of the Fittest, especially when N is large. More agents being placed in one row in the initialization will be conductive to that.*

### 4.2 Random-p, the probability of randomized move – Chaos of System

**Experiment 4:** N=(100,600),RowNum=3,RunNum=3,MaxRandom-p=2,MaxLeast-P=90,MaxCoop-P=20
**Experiment 3:** N=(100,600),RunNum=3,RowNum=3,MaxRandom-p=1,MaxLeast-P=90,MaxCoop-P=50

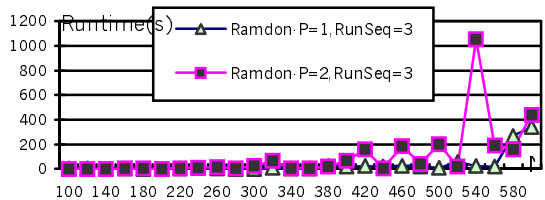

**Figure 8**. A comparison of runtime for Experiments 3 and 4.

The best runtime performances (RunSeq=3) of these two experiments are presented in Fig.8. Obviously, efficiency of Experiment 3 is much better than that of Experiment 4. However, if we set MaxRandom-p to *zero*, sometimes system cannot find any solution.

**Observation 2**: *Randomized movement is indispensable in this system. However,*

*it will bring system to chaos, if the possibility of Randomized-move is too high. So the value of Random-p should be set within a reasonable range.*

### 4.3 Least-P vs. Coop-P

In order to focus on the issue of how Least-P and Coop-P affect the system, in the following experiments, RowNum is set to 1, and all agent are assigned the same Least-P and Coop-P in initialization. Four pairs of experiments (each repeats twice) as in Table 2 are conducted to study the effect of Coop-P and Least-P under 4 different *N*. It shows that the tendency of '*runtime going shorter while Least-P and Coop-P going larger*', especially when *N* is large. This may be expressed as follows:

$$N\uparrow \Rightarrow (\text{Least-P}\uparrow \& \text{Coop-P}\uparrow \Rightarrow \text{runtime}\downarrow)\uparrow.$$

**Table 2.** Linear-regress analysis for runtime of RunSeq=1, RunNum=1, RowNum=1, MaxRandom-p=1

|   | N    | Least-P =(20,200) / Coop-P =150         | Coop-P =(20,200) / Least-P =150         |
|---|------|------------------------------------------|------------------------------------------|
| 5 | 300  | Runtime=-0.003397*Least-P+3.35567        | Runtime=-0.00082*Coop-P+3.14215          |
| 6 | 400  | Runtime=-0.022242*Least-P+7.04474        | Runtime=-0.008034*Coop-P+5.26072         |
| 7 | 700  | Runtime=-0.10868*Least-P+37.1561         | Runtime=-0.114108*Coop-P+38.7139         |
| 8 | 1000 | Runtime =-1.13953*Least-P+237.925        | Runtime=0.983722*Coop-P+209.691          |

We further investigate the effect of Coop-P and Least-P by collecting averaged runtime over the X-axis of (Coop-P+Least-P). The minimal runtime value for *N=300, 400, 700, 1000* is located respectively at (Coop-P + Least-P) = *240, 250*, 320, 350. This shows the tendency of:

$$N\uparrow \Rightarrow ((\text{Least-P+Coop-P})\uparrow \Rightarrow \text{runtime}\downarrow)\uparrow.$$

**Table 3.** Averaged runtime of two groups experiments

|       | N   | Coop-P | Least-P | Time(s) |
|-------|-----|--------|---------|---------|
| 9(1)  | 300 | 200    | 50      | 3.4     |
| 9(2)  | 300 | 125    | 125     | 4.5     |
| 9(3)  | 300 | 50     | 200     | 1.7     |
| 10(1) | 700 | 200    | 50      | 61.5    |
| 10(2) | 700 | 125    | 125     | 63.5    |
| 10(3) | 700 | 50     | 200     | 36.5    |

Another phenomenon of interest, Least-P always playing a more important role than Coop-P, is presented by experiment 9 and experiment 10 (each repeats 4 times and get the averaged values), we set RowNum=1, RunNum=1, MaxRandom-p =1 while maintaining

**Observation 3:** *When N is increasing, Least-P and Coop-P should be increased accordingly. And Least-P plays a more important role than Coop-P in systems performance.*

## 5  An ALife-Based Agent Model for General CSPs

Like *N*-queen problem, the benchmark of CSPs, a more general CSP can be defined as ***P***=(***X***, ***D***, ***R***). There is an *n*-variable set ***X***={$X_1$, $X_2$, …, $X_n$}, a domain set for each variables ***D***={$D_1$,$D_2$,…,$D_n$}, where $X_m \in D_m$, and a constraints set ***R***={$R_1$,$R_2$,…,$R_m$}, where $R_j$ is the constraint between two or more variables. A *solution* to this problem is an assignment of values {$a_1$,$a_2$,…,$a_n$} to {$X_1$,$X_2$,…,$X_n$ }, where $a_i \in D_i$ and satisfies all the constraints in ***R***. The model of an ALife-based agent system for solving such CSPs is given below, and a corresponding algorithm is presented in Fig.9.

```
1. Initialization of system;
2. Initialize the Space and agent group Φ; Timer = 0;
3. Randomly place agents to their corresponding row;
4. While not (current-space = no-solution-state) do
     For all φ∈ Φ  do /* schedule in time t (t = timer) */
       Select behavior among (Random-Move, Least-Move, Coop-Move), get new position (x', y);
       If current-position(x,y)=(x',y) then stay
       else /* move to other position */
         if Other agent φ' in position(x', y) and eat-condition = true  then eat(φ,φ');
         Update φ.energy ;Update energy of all agents that collide with φ.  Update    Space;
         If current-space= solution-state GoTo 6
       Timer ++;
5. Output "Fail to find a solution!"; GoTo 2; /* when current space = no-solution-state */
6. Output solution; If want-to-find-another-solution-with-the-current-agent-group then GoTo 3;
```

**Figure 9.** An ALife-based algorithm for CSPs.  Space complexity is $O(\Sigma |D_i|)$.

***P*** $\rightarrow$ **ALife system.**

***X*** $\rightarrow$ **Agents**. Variable can be substituted with one or more agents, and the corresponding domain can be mapped to the possible space positions of the agents. For example, agent *φ1* and *φ2* are regarded as *Xi*, then *φ1* and *φ2* can 'move' within $D_i$. If both of move to position of *b*, $b \in D_i$, it can be said that $Xi = b$.

***R* & assignment of *X*** $\rightarrow$ **Space**. It has |D| rows and |Dj| units for row *j*, total $\Sigma$|Di| units recording the number of current collisions — dissatisfactory constraints, and the current position of each agent. When an agent moves to a new position, it will lose *m* units of energy if the new assignment to the related variable dissatisfies totally *m* constrains, and the other agents corresponding to those dissatisfactory constraints will also lose energy. It will eat other agents in the new position if the energy difference is greater than a threshold. Moreover, the space will update the collision number and the agent position.

**Solution** $\rightarrow$ positions of the current survival agents. When there is no collision and only one position occupied by agents per row, a solution is found.

## 6  Conclusion

This paper described an approach of using an ALife-based distributed agent system to solve *N*-queen problems. While giving the detailed ALife-based agent model and the underlying algorithm, this paper also presented several experimental results to demonstrate how the natural rule of *Survival of the Fittest* enables the optimization of a CSP system —   it can solve a 1500-queen problem in few hundreds of seconds, and discussed the effects of behavioral parameters on the agent performance. Two features of the proposed approach can be noted as follows:

(a) The process of finding a solution is entirely determined by the locality of the individual agents. It is distributed and has no central controller to direct the process. To programmers, what they need to think about is the strategies of a single agent, but not the whole process of solving the problem;

(b) The agent's movements such as Randomized-Move, Least-Move, and Coop-Move are dynamically selected. The system dynamically evolves by eliminating bad-strategy agents through selection and contest, according to the rule of *Survival of the Fittest*.

   The proposed approach will have significant impact on solving general CSPs. An ALife-based model has been given in the paper, which provides the researchers in CSPs area a new way of studying and solving problems. Future studies based on our present work include: (i) to solve much *larger-scale N*-queen problems; (ii) to add more evolvable strategies to agents; (iii) to solve *(N+m)*-Queen problem based on an agent group evolved in solving *N*-queen problems; this will reduce the CSP solving time for the agent group; (iv) to introduce *reproduction* and *mutation* mechanisms to the system; and (v) to experimentally investigate the ALife-based algorithm in solving other CSPs.